\documentclass{article}
\usepackage[final,nonatbib]{neurips_2018}
\usepackage{wrapfig}
\let\labelindent\relax
\usepackage{enumitem,soul}
\usepackage{rotating}
\usepackage{chngcntr}
\usepackage{apptools}
\usepackage{listings}
\usepackage{graphicx}
\usepackage{graphics}
\usepackage{amssymb}
\usepackage{amsmath}
\usepackage{bbm}
\usepackage{color,cite}
\usepackage{soul}
\usepackage{xspace}
\usepackage{algpseudocode}
\usepackage{algorithm,algorithmicx}
\usepackage{algcompatible}
\usepackage{bbm}
\usepackage{comment}
\usepackage{caption}
\usepackage{subcaption}
\usepackage{psfrag}
\usepackage{pgfplots}
\usepackage{rotating}
\usepackage{graphicx}
\usepackage{sidecap}
\usepackage{diagbox}
\usepackage{adjustbox}
\usepackage{multirow}
\usepackage{tikz}
\usetikzlibrary{math}
\usetikzlibrary{shapes,arrows}
\usetikzlibrary{arrows.meta}
\tikzset{>={Latex[width=2.5mm,length=2.5mm]}}
\usetikzlibrary{positioning,calc}
\tikzstyle{block}=[draw opacity=0.7,line width=1.4cm]
\usetikzlibrary{decorations.pathreplacing,decorations.markings}
\usetikzlibrary{matrix,chains,positioning,decorations.pathreplacing,arrows}
\usetikzlibrary{positioning,calc}

\tikzset{block/.style={%
        inner xsep=1mm,
        inner ysep=1.5mm,
        rectangle,very thick,draw}}
\tikzset{sum/.style={%
        circle,
        minimum size=2mm,inner xsep=1.2mm,inner ysep=1.2mm,
        very thick,draw}}
\tikzset{point/.style={%
        minimum size=0mm,inner xsep=4mm,inner ysep=0mm,draw}}
\tikzset{link/.style={->,very thick,>=stealth}}
\tikzset{undirlink/.style={<->,very thick,>=stealth}}
\tikzset{pole/.style={cross out, draw=black, minimum size=2*(#1-\pgflinewidth), inner sep=0pt, outer sep=0pt},cross/.default={1pt}}

\tikzset{%
  every neuron/.style={
    circle,
    draw,
    minimum size=0.2cm
  },
  neuron missing/.style={
    draw=none, 
    scale=1,
    text height=0.2cm,
    execute at begin node=\color{black}$\vdots$
  },
}

\title{\LARGE \bf
Learning Contraction Policies from Offline Data}

\author{{Navid Rezazadeh}\\
\textit{Mechanical and Aerospace Eng.} \\
\textit{UCI}
\and
{~Maxwell Kolarich~}\\
\textit{Aerospace Eng.} \\
\textit{UIUC}
\and
{~Solmaz S. Kia, \emph{IEEE member, Senior}}\\
\textit{Mechanical and Aerospace Eng.} \\
\textit{UCI}
\and 
{Negar Mehr}\\
\textit{Aerospace Eng.} \\
\textit{UIUC}}

\author{%
  Navid Rezazadeh\\
  Mechanical and Aerospace Eng. Dept.\\
  UCI\\
  \texttt{nrezazad@uci.edu}
  \And
  Maxwell Kolarich\\
  Aerospace Eng. Dept.\\
  UIUC\\
  \texttt{mak13@illinois.edu }
  \And
  Solmaz S. Kia\\
  Mechanical and Aerospace Eng. Dept.\\
  UCI\\
  \texttt{solmaz@uci.edu}
  \And
  Negar Mehr\\
  Aerospace Eng. Dept.\\
  UIUC\\
  \texttt{negar@illinois.edu}
}

\newcommand{\real}{{\mathbb{R}}}

\newcommand{\Lnorm}{\left\|} 
\newcommand{\Rnorm}{\right\|}

\newcommand{\vect}[1]{\boldsymbol{\mathbf{#1}}}

\newcommand{\dvect}[1]{\dot{\vect{#1}}}

\newcommand{\SUM}[2]{\sum_{#1}^{#2}}
 \newcommand{\boxend}{\hfill \ensuremath{\Box}}

\newcommand{\x}{\vect{x}}
\newcommand{\xt}{\x_t}

\newcommand{\xtpl}{\x_{t+1}}
\newcommand{\xtplest}{{\x}^{'}_{t+1}}

\newcommand{\xpr}{\tilde{\x}}

\newcommand{\xtpr}{\xpr_t}

\newcommand{\xtplpr}{\xpr_{t+1}}
\newcommand{\xtplprest}{\xtplpr^{'}}

\newcommand{\xr}{\x^r}
\newcommand{\xdot}{\dvect{x}}
\newcommand{\xest}{{\x}^{'}}

\newcommand{\xesttpl}{\xest_{t+1}}

\newcommand{\dx}{\delta \vect{x}}
\newcommand{\dxt}{\dx_t}
\newcommand{\dxtpl}{\dx_{t+1}}

\newcommand{\Dx}{\Delta \vect{x}}
\newcommand{\Dxt}{\Dx_t}

\newcommand{\uu}{\vect{u}}
\newcommand{\ut}{\uu_t}

\newcommand{\ux}{\uu(\x)}
\newcommand{\uxt}{\uu(\xt)}

\newcommand{\xu}{\begin{bmatrix}\x\\\uu\end{bmatrix}}
\newcommand{\xupr}{\begin{bmatrix}\xt\\\ut\end{bmatrix}}
\newcommand{\xux}{\begin{bmatrix}\x\\\ux\end{bmatrix}}
\newcommand{\xuxpr}{\begin{bmatrix}\xpr\\\uxpr\end{bmatrix}}

\newcommand{\T}{\vect{\Theta}}

\newcommand{\M}{\vect{M}}

\newcommand{\D}{\mathcal{D}}
\newcommand{\Dp}{\mathcal{D}'}

\newcommand{\fest}{{f}^{'}}

\newcommand{\Test}{\hat{\T}}
\newcommand{\uest}{\hat{\uu}}

\newcommand{\Lfest}{\mathsf{L}_{f^{'}}}
\newcommand{\Lffest}{\mathsf{L}_{f-\fest}}
\newcommand{\Lux}{\mathsf{L}_{\uu}}
\newcommand{\Lfestu}{\mathsf{L}_{f^{'}_u}}
\newcommand{\LTij}{\mathsf{L}_{\T_{ij}}}
\newcommand{\LC}{\mathsf{L}_{C}}

\newcommand{\wf}{\vect{w}_f}
\newcommand{\wT}{\vect{w}_{\Theta}}

\newcommand{\wU}{\vect{w}_{\uu}}

\newcommand{\wUopt}{\wU^\star}
\newcommand{\wTopt}{\wT^\star}

\newcommand{\EX}{\mathbb{E}}
\newcommand{\X}{\mathcal{X}}
\newcommand{\UU}{\mathcal{U}}
\newcommand{\XTU}{\X \times \UU}

\newcommand{\rev}[1]{{\color{black}#1}}

\newtheorem{prop}{Proposition}

\newtheorem{lem}{Lemma}
\newtheorem{defn}{Definition}

\newenvironment{proof}[1][Proof]{\begin{trivlist}
\item[\hskip \labelsep {\bfseries #1}]}{\boxend\end{trivlist}}

\makeatletter
\renewcommand*{\@opargbegintheorem}[3]{\trivlist
      \item[\hskip \labelsep{\bfseries #1\ #2}] \textbf{(#3)}\ \itshape}
\makeatother

\makeatletter
\let\NAT@parse\undefined
\makeatother
\usepackage[colorlinks]{hyperref}

\usepackage{setspace}
\begin{document}
\maketitle
\thispagestyle{plain}
\pagestyle{plain}

\begin{abstract}  
This paper proposes a data-driven method for learning convergent control policies from offline data using Contraction theory. Contraction theory enables constructing a policy that makes the closed-loop system trajectories inherently convergent towards a unique trajectory. At the technical level, identifying the contraction metric, which is the distance metric with respect to which a robot's trajectories exhibit contraction is often non-trivial. We propose to jointly learn the control policy and its corresponding contraction metric while enforcing contraction. To achieve this, we learn an implicit dynamics model of the robotic system from an offline data set consisting of the robot's state and input trajectories. Using this learned dynamics model, we propose a data augmentation algorithm for learning contraction policies. We randomly generate samples in the state-space and propagate them forward in time through the learned dynamics model to generate auxiliary sample trajectories. We then learn both the control policy and the contraction metric such that the distance between the trajectories from the offline data set and our generated auxiliary sample trajectories decreases over time. We evaluate the performance of our proposed framework on simulated robotic goal-reaching tasks and demonstrate that enforcing contraction results in faster convergence and greater robustness of the learned policy.

\end{abstract}

\section{Introduction}

While learning-based controllers have achieved significant success, they still lack safety guarantees. For instance, in general, the temporal evolution of a robot's trajectories under a learned policy cannot be certified. On the other hand, when a system's dynamics are known, control-theoretic properties, such as stability and contraction, directly examine the temporal progression of a system's states to verify whether a system remains within a safe set, and whether the system's trajectories converge. In this paper, we seek to enforce the desired temporal evolution of the closed-loop system's states while learning the policy from an offline set of data, i.e.\ we seek to learn control policies such that under the learned policy, the convergence of a robot's trajectories is achieved.

To achieve such trajectory convergence, our design approach leverages Contraction theory~\cite{lohmiller1998contraction}. Contraction theory provides a framework for identifying the class of nonlinear dynamic systems that have asymptotic convergent trajectories. Intuitively, a region of the state space is a contraction space if the distance between any two close neighboring trajectories decays over time. This notion of convergence is relevant to many robotic tasks such as tracking controllers where we want a robot to either reach a goal or track a reference trajectory. In this paper, we want to learn policies from offline data such that they achieve convergence of a robot's trajectories in closed loop. While contraction theory provides a simple and intuitive characterization of convergent trajectories, finding the distance metric with respect to which a robot's trajectories exhibit contraction -- which is called the \emph{contraction metric} -- is often non-trivial. To address this challenge, we propose to \emph{jointly} learn the robot policy and its corresponding contraction metric.

\begin{figure}
    \centering
    \includegraphics[width=.8\textwidth]{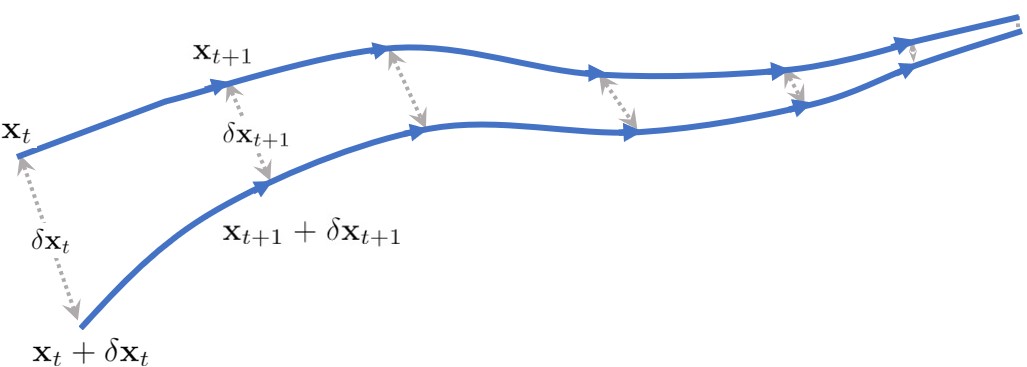}
    \caption{{\small The schematic of two neighboring trajectories that exhibit contraction. The distance between the trajectories decreases over time: $\Lnorm\dxtpl\Rnorm < \Lnorm\dxt\Rnorm$, i.e. trajectories converge. }
    }
    \label{fig:contraction}
\end{figure}

We learn the robot dynamics model from an offline data set consisting of the robot's state and input trajectories. This setting is similar to the setting of offline model-based reinforcement learning (RL) where a dynamics model and a policy are learned from a set of robot trajectories that are collected offline. Learning from offline data is appropriate for safety-critical applications where online data collection is dangerous~\cite{levine2020offline}. We learn a dynamics model of the system from the data and propose a data augmentation algorithm for learning contraction policies. 
Randomly sampled states are propagated forward in time through the learned dynamics model to generate auxiliary sample trajectories.
\rev{We then learn both our policy and our contraction metric such that the distance between the robot trajectories from the data set and the auxiliary sample trajectories decreases over time. Learning contraction policies is particularly relevant to offline RL as it allows us to regard the errors in the learned dynamics model as external disturbances and obtain a tracking error bound in regions where the learning errors of the dynamics model are bounded~\cite{NMB-ST-NM-JJES-VS:20,HT-SJC-JJES:21}.}

We evaluate the performance of our proposed framework on a set of simulated robotic goal-reaching tasks. 
The performance of our proposed framework is compared with a number of control algorithms. We demonstrate that as a result of enforcing contraction, the robot's trajectories converge faster to the goal position with a higher degree of accuracy. It is further shown that learning contraction policies increases the robustness of the learned policy with respect to learned dynamics model mismatch, i.e.\ enforcing contraction increases the robustness of the learned policies. In summary, our contributions are the following:
\begin{itemize}
\item We propose a framework for learning convergent robot policies from an offline data set using Contraction theory.
\item We develop a data augmentation algorithm for learning contraction policies from the offline data set.
\rev{
\item We provide a formal analysis for bounding contraction policy performance as a function of dynamics model mismatch.
}
\item We perform numerical evaluations of our proposed policy learning framework and demonstrate that enforcing contraction results in favorable convergence and robustness performance.
\end{itemize}

The organization of this paper is as follows. In Section~\ref{sec:related}, we provide an overview of the related and prior work. We provide an overview of Contraction theory in Section~\ref{sec:contraction} and present our problem formulation in Section~\ref{sec:problem-form}. We then discuss our proposed framework in Section~\ref{sec:learning-contraction}. \rev{Section~\ref{sec:robustness} provides a discussion and analysis of the robustness of learned contraction policies.} In Section~\ref{sec:implementation}, we evaluate and compare the performance of our policy learning algorithm. Finally, we will conclude the paper in Section~\ref{sec:conclusion}.

\section{Related Work}\label{sec:related}

For systems with unknown dynamics, several offline RL algorithms have been developed recently which either directly learn a policy using an offline data-set~\cite{TH-AZ-PA-SL:18,AK-AZ-GT-SL:20,TY-GT-LY-SE-JYZ-SL-CF-TM:20,SL-AK-GT-JF:20} or learn a surrogate dynamics model from the offline data to learn an appropriate policy~\cite{LK-MB-SL:19,TMM-JB-MCJ:20}. However, the majority of such RL algorithms lack formal safety guarantees, and the convergent behavior of the learned policies is not certified~\cite{sun2020learning, berkenkamp2017safe}. 

When the system dynamics are known, robust and certifiable control policy design can be achieved through various control-theoretic methods such as reachability analysis~\cite{SV-SK-HL-FB-JW-SW-MJ-RV:19}, Funnels~\cite{RT:09,AM-RT:17}, and Hamilton-Jacobi analysis~\cite{SLH-MC-SH-SB-JFF-CJT:17,SB-MC-JFF-CJT:17}. Lyapunov stability criteria, Contraction Theory, and Control Barrier Functions have been extensively utilized for providing strong convergence guarantees for nonlinear dynamical systems~\cite{HKK-JWG:02,JC-FC-CJT-KS:20,AJT-AS-YY-ADA:20,AMZ-AME-MEL-FS:21,lohmiller1998contraction}. However, even when the dynamics are known, finding a proper Lyapunov function or a control barrier function is itself a challenging task. To address these challenges, learning algorithms have been utilized for learning the Lyapunov and Control Barrier Functions~\cite{SMR-FB-AK:18,Ar-HH-LL-HZ-DVD-ST-NM:20,SC-MF-MM-GJP-VMP:21}. ~\cite{sun2020learning} and \cite{tsukamoto2021learning} propose to learn contraction metrics to find contraction control policies for known systems. 

\rev{
Various recent works have considered combining control-theoretic tools with learning algorithms to enable learning safe policies even when dynamics are unknown. ~\cite{SMK-AB:11,JU-SH:17} consider learning stable dynamics models. In~\cite{SSV-VS-JJES-MP:18}, Contraction theory is used to learn stabilizable dynamics models of unknown systems. In~\cite{berkenkamp2017safe,AJT-VDV-HML-YY-ADA:19}, Lyapunov functions are used for ensuring the stability of the learned policies.~\cite{JZK-GM:19} proposed to learn the system dynamics and its corresponding Lyapunov function jointly to ensure the stability of the learned dynamics model.
}



\rev{In this work, we consider learning contraction policies from offline data for systems with unknown dynamics.} Our work is closely related to~\cite{sun2020learning} and \cite{tsukamoto2021learning}, where Contraction theory has been used for certifying convergent trajectories. 
The current work is different in that, unlike these approaches where dynamics are explicitly known and assumed to be control-affine, \rev{we consider access to only an offline data set. We assume that we can learn an implicit model of system dynamics, in the form of a neural network function approximator, and provide robustness guarantees with respect to the errors of the learned dynamics model.}

\section{Contraction Theory}\label{sec:contraction}

Contraction theory assesses the stability properties of dynamical systems by studying the convergence behavior of neighboring trajectories~\cite{lohmiller1998contraction}. The convergence is established by directly examining the evolution of the weighted Euclidean distance of close neighboring trajectories.
\rev{
Formally, consider a differentiable autonomous discrete-time dynamical system $g(\x): \real^n \to \real^n$ defined as
\begin{align}\label{eq::autoDynamics}
    \xtpl=g(\xt), 
\end{align} 
with Jacobian
\begin{align}
    \nabla g(\xt) = \left. \frac{\partial g(\x)}{\partial \x } \right\vert_{\x=\xt}.
\end{align} 
Now, consider a differential displacement $\dxt$. The differential displacement dynamics at $\xt$ are governed by
\begin{align}\label{eq::virtualDynamics}
    \dxtpl=\nabla g(\xt)\dxt. 
\end{align}
The system dynamics $g(\xt)$ are contractive if there exists a full rank state dependant metric $\T(\x) \in \real^n \times \real^n$ such that the system trajectories satisfy 
\begin{align}\label{eq::contDef}
\| \T(\xt)\dxt \| > \| \T(\xtpl)\dxtpl \|.
\end{align}
}
Equation~\eqref{eq::contDef} indicates that the weighted distance between any two infinitesimally close states decreases as the dynamics evolve~\cite{WL-JJES:98}. When $\T(\x)=\vect{I}$ the distances between trajectories are measured in the Euclidean norm sense. Figure~\ref{fig:contraction} illustrates the behavior of two trajectories of a contractive system when $\T(\x)=\vect{I}$. 

\rev{
For a small finite displacement $\Dxt$, as an approximation of infinitesimal small displacement $\dxt$, the first-order Taylor expansion of the system dynamics allows us to locally approximate the forward evolution of the displacement
\begin{align}
    \nabla g(\xt) \Dxt \approx g(\xt + \Dxt) - g(\xt).
\end{align}
Thus, we may approximate the contraction condition~\eqref{eq::contDef} as
\begin{align}\label{eq::contCond}
\| \T(\xtpl)\left(g(\xt + \Dxt) \!- \!g(\xt) \right)\|\!-\!\| \T(\xt)\Dxt \|\!\!<\! 0.
\end{align}

Establishing a system as contractive allows for several useful stability properties to be deduced. We state motivating results from~\cite{lohmiller1998contraction} in the following definition and proposition.
\begin{defn}
    Given the discrete-time system $\xtpl=g(\xt)$, a region of the state space is called a contraction region with respect to a uniformly positive definite metric $\M(\xt) = \T(\xt)^T\T(\xt)$, if in that region
    \begin{align}\label{eq::contDef2}
          \nabla g(\xt)^T \M(\xtpl) \nabla g(\xt) - \M(\xt) < 0,
    \end{align}
\end{defn}
 
\begin{prop} \label{prop::contRegion}
    A convex contraction region contains \textit{at most one equilibrium point}.
\end{prop}

It is shown in~\cite{lohmiller1998contraction} that~\eqref{eq::contDef2} is equivalent to condition~\eqref{eq::contDef} holding for all $\xt$ in the contraction region.
Thus, by Proposition~\ref{prop::contRegion}, we may conclude that a unique equilibrium exists within a convex region if~\eqref{eq::contDef} holds everywhere inside the region. Therefore,~\eqref{eq::contCond} represents a useful numerical analog that can be enforced in order to drive a region towards being contractive. By choosing a  set of $\Dxt$, we will use condition~\eqref{eq::contCond} to enforce contracting behavior of the closed-loop system.
}
Going beyond autonomous systems, when a system is subject to control input $\ut$, i.e., $\xtpl=g(\xt)=f(\xt,\ut)$, contraction theory can be used to design state feedback policies $\ut=\uxt$ such that the closed-loop system trajectories converge to a given reference state. \rev{This may be done by determining $\uxt$ such that the convex region of interest is contractive and the unique equilibrium is the desired reference state. Such a control design process is outlined in the following sections.}
\section{Problem Formulation}\label{sec:problem-form}
We consider the problem of control policy design for a robot with unknown discrete-time dynamics model $f(\x ,\uu): \XTU \to \X$, where $\X \in \real^n$ \rev{is convex}, $\UU \in \real^m$. We assume that we can use an offline data set $\D$ consisting of tuples of state transitions and control inputs $(\xt,\xtpl,\ut)$ satisfying unknown system dynamics
\begin{align}\label{eq::trueDynamics}
    \xtpl=f(\xt,\ut).
\end{align}

Our objective is to obtain a data-driven state-feedback control policy $\ut = \uxt$ to steer the system~\eqref{eq::trueDynamics} towards a desired reference state $\xr \in \real^n$, i.e. $\xt\to\xr$ as $t\to\infty$. 
\rev{To compensate for the lack of knowledge of the true system dynamics, we propose using a model of the system dynamics that we learn from the offline data $\D$. Note that this indicates that our learned dynamics model may still not be available explicitly and may only be available as implicit dynamics such as neural networks approximators.} More specifically, we aim to design a control policy $\uxt$ that leverages the learned dynamics model
\begin{align}\label{eq::modelDynamics}
    \xesttpl=\fest(\xt,\ut), 
\end{align}
to drive the system asymptotically to $\xr$. 


\section{Learning Deep Contraction Policies}\label{sec:learning-contraction}
\begin{figure}
    \centering
    \includegraphics[width=.5\textwidth]{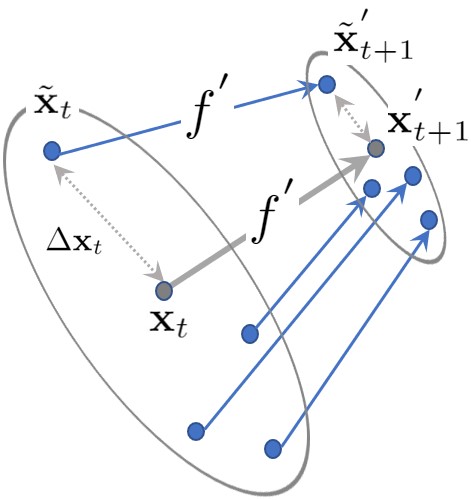}
    \caption{{\small \rev{We sample a small displacement $\Dxt$ around the data point $\xt$ to augment an auxiliary point $\xtpr = \xt+\Dxt$ to our data set. Then, we propagate the auxiliary state $\xtpr$ and the actual state $\xt$ through our learned dynamics model $\fest$ under feedback control law $\uxt$ to calculate the next states: $\xtplprest$ and $\xtplest$, respectively. Finally, we require that the weighted distance between the two states decreases over time as stated in condition~\eqref{eq::cont_notlearned}.}}}
    \label{fig:sampling}
\end{figure}

To develop a policy that results in contractive behavior, we seek to enforce the approximate condition in~\eqref{eq::contCond}, requiring the weighted distances of close neighboring trajectories to decrease over time. To enforce this condition, we need to ensure that we have sufficiently close neighboring points for each point within our training set. \rev{However, our training data set may not include such neighboring trajectories. We augment our data set with auxiliary trajectories that enable us to enforce this condition at each data point. That is, for each $\xt \in \D$, we augment our data set with a $\Dxt$ sampled from
\begin{align}
    \label{eq::omega}
    \Delta_t=\left\{\Dx_t \in\real^n\,\Big|\, \|\Dx_t\|<\lambda \right\},
\end{align}
where the parameter $\lambda$ is set in the training process. We sample points from $\Delta_t$ to ensure that $\Dxt$ is a small displacement with respect to the training data set. Then, for each data point $\xt$, we create the auxiliary state $\xtpr = \xt + \Dxt$. Both of these points are propagated through our learned dynamics model to calculate the states at the next time step: $\xtplest = \fest(\xt,\uxt)$ and $\xtplprest = \fest(\xtpr,\uu(\xtpr))$. The initial state, the auxiliary state, and the predicted evolution of these two states are then combined into a tuple $(\xt,\xtpr,\xtplest,\xtplprest)$. The collection of all such tuples over each $\xt \in \D$ form the augmented data set $\Dp$.}

Now, we want to evaluate the contracting behavior of the controller $\uxt$ through the learned model on the augmented data set $\Dp$. Thus, we seek to enforce condition~\eqref{eq::contCond} for the elements of $\Dp$
\begin{align}\label{eq::cont_notlearned}
    \| \T(\xtplest)(\xtplprest - \xtplest)\|-\| \T(\xt)\,\Dxt \|<0,
\end{align}
with respect to a contraction metric $\T(\xt)$. Contractive behavior is illustrated in Figure~\ref{fig:sampling}, showing the weighted distance between $\xtpr$ and $\xt$ decays as the system evolves to $\xtplprest$ and $\xtplest$. We evaluate the approximate contraction condition only at the states $\xt$ that exist in the data set $\D$. \rev{This is due to the fact that the dynamics model is learned from $\D$ and hence $\fest$ is expected to behave the most accurately at these points, which in turn will increase the quality of the learned policy. This will enforce contractive behavior with respect to the learned dynamics model $f'$. Later we will discuss how we can ensure contractive behavior of the closed-loop behavior of the true dynamics model $f$.}



Since in general, the contraction metric $\T(\x)$ is not known, and it is directly coupled to the control policy, we propose a learning-based approach to jointly learn both the control policy and the metric with respect to which the policy exhibits contraction. We refer to such a policy as a deep contraction policy. To this end, 
let us start by assuming that we know a control policy $\ux$ that makes $\fest(\x,\ux)$ contractive. Consider now that we want to learn a corresponding contraction metric. Let this contraction metric be represented by a model $\Test(\x;\wT)$ which is parameterized by weights $\wT$. We then obtain the best parameters of this contraction metric, denoted by $\wTopt$, from
\begin{align}
    \wTopt = \underset{\wT}{\textup{argmin}}\,\,\, L_{\T}(\D';\wT),
\end{align}
where
\begin{align}\label{eq:loss-contraction}
     L_{\T}(\D';\wT) = \EX_{\Dp}\Big(\| \Test(\xtplest;\wT)(\xtplprest - \xtplest) \| -\| \Test(\xt;\wT) \Dxt \|\Big). \nonumber
\end{align}
The term $\| \Test(\xtplest;\wT)(\xtplprest - \xtplest) \|-\| \Test(\xt;\wT) \Dxt \|$ is an approximate measure of the contraction condition~\eqref{eq::cont_notlearned} which ideally should be negative for all elements of $\Dp$. \rev{Since enforcing~\eqref{eq::cont_notlearned} directly results in a non-differentiable optimization, we minimize~\eqref{eq:loss-contraction} as a proxy for~\eqref{eq::cont_notlearned}. Note that $L_{\T}$ is computed over all data points in $\Dp$. When paired with differentiable contraction metric $\Test(\x;\wT)$, the choice of loss function~\eqref{eq:loss-contraction} is differentiable and is amenable to gradient decent optimization.}


Now, let's consider the more general case where both the policy and its contraction metric are unknown. We want to learn both the state-feedback policy and the contraction metric together. We want to learn a control policy represented by a function approximator $\uest(\x;\wU)$, parameterized by weights $\wU$, such that the closed-loop system is contractive with respect to the metric model $\Test$. To achieve this, we propagate the initial data points in $\Dp$ with the control policy model $\uest(\x;\wU)$ as $\xtplest = \fest(\xt,\uest(\xt;\wU))$ and $\xtplprest = \fest(\xtpr,\uest(\xtpr;\wU))$.

We obtain the parameters of the contraction metric $\wT$, denoted by $\wTopt$, and the parameters of the control policy $\wU$, denoted by $\wUopt$, by minimizing a loss function $L_u$ over the data set $\D'$
\begin{align}
    (\wUopt,\wTopt) = \underset{\wU,\wT}{\textup{argmin}}\,\,\,L_{u}(\D';\wU,\wT),
\end{align}
where
{\begin{align}\label{eq::controller_loss}
     L_{u}(\D';\wU,\wT) = 
      \EX_{\D'} \Big(\|  \Test(\xtplest;\wT)(\xtplpr-\xtplest)) \| -\| \Test(\xt;\wT) \Dxt \| \Big). \nonumber
\end{align}}

Loss function~\eqref{eq::controller_loss} ensures that the region of interest $\X$ is contractive with respect to $\Test$ and the learned dynamics model $f'$. However, so far there has been no mechanism to ensure that the unique equilibrium of the contractive system is indeed the desired reference value $\xr$. To alleviate this, we need the learning process to be aware of the desired reference value, which we would like to be the equilibrium of the contraction region. The measure of awareness that we introduce is based on the ability of the controller $\uest(\x;\wU)$ to steer the system from an initial state $\x_0 \in \mathcal{X}$ to the desired state value $\xr$ in $k$ time steps, i.e. how close $\xest_k$ gets to $\xr$. Therefore, to enforce the system's states to contract to $\xr$, we add another penalty term to our loss function to obtain the final loss function utilized for learning the policy and contraction metric:
\begin{align}
    L&(\D',\mathcal{Y};\wT,\wU) \!\!=\!\! L_{u}(\D';\wT,\wU) \!+\! \alpha L_{\textup{tr}}(\mathcal{Y};\wU),
\end{align}
where 
\begin{align}
    L_{\textup{tr}}(\mathcal{Y};\wU) = \SUM{\x_0 \in \mathcal{Y}}{} \Lnorm \xest_{k}(\x_0) - \xr \Rnorm,
\end{align} is the tracking loss with $\alpha\in\real_{>0}$ as the penalty factor. Here, $\xest_k(\x_0)$ is the $k^{\text{th}}$ state value of the process $\xest_{t+1} = \fest(\xest_{t},\uest(\xest_{t};\wU))$, initialized at $\xest_{0}=\vect{x}_0$ where $\vect{x}_0$ is drawn from a countable set $\mathcal{Y} \in \mathcal{X}$. The number of time steps $k$ is set by the designer and, as the reader may infer, affects the transient behavior of the closed-loop system.
\begin{algorithm}[t]
\caption{Learning Deep Contraction Policies}
\label{alg:deep_contraction}
{
\begin{algorithmic}[1]
\STATE ${\textbf{Input:}}$
\STATE \hspace{15pt} $\textup{Data set: }(\xt,\xtpl,\ut) \in \D$
\STATE \hspace{15pt} $\rev{\textup{Set of sampled initial states}: \x_0 \in \mathcal{Y}}$
\STATE \hspace{15pt} $\textup{Reference state: }\xr$
\STATE ${\textbf{Init:}}$
\STATE \hspace{15pt}$\fest(\xt,\ut) \gets \textup{learned dynamics using } \D$
\STATE \hspace{15pt}$\wT,\wU \gets \textup{randomly sampled}$
\FOR{$\,\, N_{\text{epochs}}$}
\STATE $\textup{Calculate } \xest_k \textup{'s using } \mathcal{Y} \textup{ and } \fest(\xt,\uu(\xt;\wf))$
\STATE $\textup{Calculate } L_{\textup{tr}}(\wU) \textup{ using } \x_0 \in \mathcal{Y} \textup{ and } \x_k\textup{'s}$
\STATE $\Dxt \gets \textup{ uniform random sample from } \Delta_t$
\STATE $\textup{Create } \Dp \textup{ using sampled } \Dxt$
\STATE $\textup{Calculate } L_{u}(\wT,\wU) \textup{ using } \Dxt \textup{'s} \textup{ and } \Dp$
\STATE $L(\wT,\wU) \gets L_{u}(\wT,\wU) + \alpha L_{\textup{tr}}(\wU)$
\STATE $\textup{Calculate gradients } \nabla_{\wT}{L} \textup{ and } \nabla_{\wU}{L}$
\STATE $\textup{Update } \wT \textup{ and } \wU$
\ENDFOR
\end{algorithmic}
}
\end{algorithm}

\section{Contraction of True Dynamics Under Learned Policy}\label{sec:robustness}
A major concern regarding control policy design using a learned model from offline data is that of model mismatch. In order to bound the controller performance degradation, we assume a known upper bound on the Lipschitz constant of the model error $f(\xt,\ut)-\fest(\xt,\ut)$, which we denote as $\Lffest$. In practice, such an upper bound may be estimated by fitting a Reverse Weibull distribution over the data set $\D$~\cite{GRW-BPZ:96,CK-GC-NO-DB:21}.

\begin{lem}\label{lem::estimation_error_bound}
Consider an unknown system $f(\x,\uu)$ and its learned model $\fest(\x,\uu)$ with an upper-bound estimation on the Lipschitz constant of $f(\x,\uu)-\fest(\x,\uu)$ as $\Lffest$. The error between the learned model and the unknown system is bounded by $\varepsilon$, i.e. $\| f(\x,\uu) - \fest(\x,\uu) \|<\varepsilon,\,\,$ for all $(\x,\uu) \in \XTU$ where
\begin{align}\label{eq::eps}
    \varepsilon = \underset{(\xt,\xtpl,\ut) \in \D}{\textit{max}} \| \xtpl - \fest(\xt,\ut) \|  + \Lffest \mathsf{D}
\end{align}
with $\mathsf{D} = \underset{(\x,\uu) \in \XTU}{\textit{max}}\,\,\underset{(\xt,\ut) \in \D}{\textit{min}}\,\, \left\| \xu - \xupr \right\|.$
\end{lem}

\noindent \ 

\begin{proof}
    See Appendix.
\end{proof}

\noindent The constant $\mathsf{D}$ in Lemma \ref{lem::estimation_error_bound} is the maximum distance that a point $(\x,\uu) \in \XTU$ can have from its nearest data point $(\xt,\ut) \in \D$.

Deep contraction policy learning proposed in Algorithm~\ref{alg:deep_contraction} ideally ensures contractive behavior of the controlled learned system $\fest(\xt,\uxt)$ at the states $\xt \in \D$. More specifically, by defining an approximate measure of contraction condition~\eqref{eq::contDef} as $C_{g(\xt)}: \mathcal{X} \times \Delta_t \to \real$
\begin{align*}
    C_{g(\xt)}(\xt,&\Dxt) =
    \|\Test(g(\xt))(g(\xtpr)-g(\x)) \|-\| \Test(\xt) \Dxt \|,
\end{align*}
the controlled learned model being contractive is equivalent to $\EX_{\Dxt} \big(C_{\fest(\xt,\uest(\xt))}(\xt,\Dxt)\big)<0$ for all $\xt \in \D$ and $\Dxt \in \Delta_t$. Hence, it remains for us to verify whether the learned policy exhibits contraction with the true unknown system dynamics in the sense of contraction condition~\eqref{eq::contDef}, i.e. $\EX_{\Dxt} \big(C_{f(\xt,\uest(\xt))}(\xt,\Dxt)\big)<0$ for all $\xt \in \mathcal{X}$ and $\Dxt \in \Delta_t$. We seek to derive a condition under which we are guaranteed that the controlled true dynamics are also contractive with the learned policy. To arrive to such quantification, we begin with contraction of the learned model $C_{\fest(\xt,\uest(\xt))}(\xt,\Dxt)$ at the training points, $\xt \in \D$ and end with an upper bound estimation of the contraction of the true dynamics $C_{f(\xt,\uest(\xt))}(\xt,\Dxt)$ at any points $\x \in \mathcal{X}$. The following Proposition establishes the condition under which the approximate contraction measure holds for the true robot dynamics under the trained $\uxt$. 
\begin{prop}
Let $\EX_{\Dxt} \big(C_{\fest(\xt,\uest(\xt))}(\xt,\Dxt)\big)<0$ for all $\xt \in \D$. Let the Lipschitz constant of $\Test_{ij}(\xt)$, $f(\xt,\ut) - \fest(\xt,\ut)$, $\fest(\xt,\ut)$, $\uest(\xt)$, $\fest(\xt,\uest(\xt))$ and $C_{\fest(\xt,\uest(\xt))}(\xt,\Dxt)$ be given as $\LTij$,$\Lffest$, $\Lfest$, $\Lux$, $\Lfestu$, and $\LC$, respectively. Additionally, let $|\Test_{ij}(\xt)|<\gamma$, $\lambda$ be given by~\eqref{eq::omega}, and $\varepsilon$ be given as in~\eqref{eq::eps}. Then the true dynamics~\eqref{eq::trueDynamics} are contractive under the trained controlled policy $\uest(\xt)$, i.e. $\EX_{\Dxt}\big(C_{f(\xt,\uest(\xt))}(\xt,\Dxt)\big)<0$ for all $\xt \in \mathcal{X}$ and $\Dxt \in \Delta_t$, if
\begin{align}
    \zeta + \lambda (\varepsilon \tau \Lfestu + (\varepsilon \tau + n\gamma) \Lffest ( 1 + \Lux))< 0,
\end{align}
where $\tau = \sqrt{\underset{ij}{\sum}\,\LTij^2}$ and $\zeta = \underset{\x \in \mathcal{X}}{\textit{max}}\,\,\underset{\xt \in \D}{\textit{min}}\,\,  C(\xt) + \LC \Lnorm \xt - \x \Rnorm$.
\end{prop}

\noindent \ 

\begin{proof}
    See Appendix.
\end{proof}

\section{Implementation \& Evaluation}\label{sec:implementation}
\begin{figure*}
    \centering
    \includegraphics[width=1\textwidth]{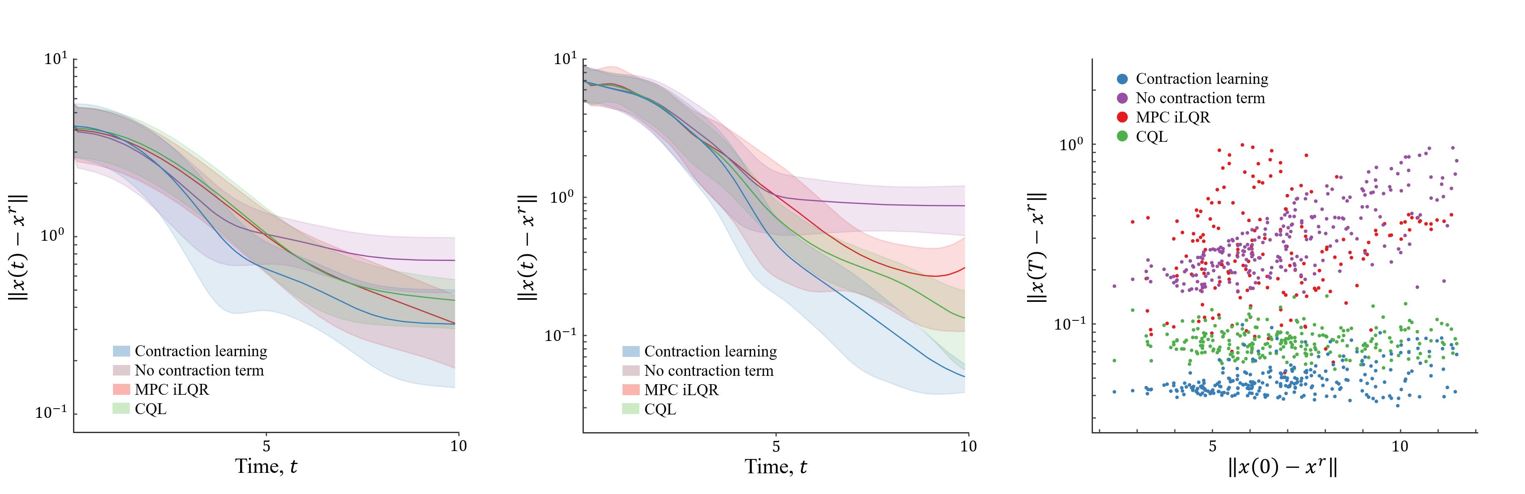}
    \caption{\small \rev{Norm of the tracking error over a collection of 256 initial states for the 2D car problem (left) and the 3D drone problem (middle). The $y$ axis is shown on a logarithmic scale and results capture the mean plus and minus one standard deviation. We see that the additional complexity of the 3D drone over the 2D car model allows for greater variation in algorithm performance. \textbf Norm of the average final tracking error versus the norm of the initial tracking error (right). As the initial states approach the boundary of the region of interest, controller performance tends to degrade.}}
    \label{fig::multiplot}
\end{figure*}

We evaluate the performance of the contraction policies in a set of goal-reaching robotic tasks by comparing our method against a number of offline control methods suitable for systems with learned dynamics models. In particular, we compare our framework with the following algorithms:
\begin{enumerate}[font=\bfseries]
  \item \textbf{MPC:} An iterative Linear Quadratic Controller (iLQR) as described in~\cite{6386025} ran in a receding horizon fashion.
  \item \textbf{Learning without contraction:} To evaluate the effectiveness of the contraction penalty, we further evaluate the robot's performance in the absence of any contraction terms in the loss function.
  \item \textbf{Reinforcement Learning:} We also use the state-of-the-art offline RL method Conservative Q-Learning (CQL)~\cite{DBLP:journals/corr/abs-2006-04779} for further comparisons.
\end{enumerate}

We evaluate the performance of our approach on two different robotic settings involving nonlinear dynamical systems of varying complexity represented by neural networks. The dynamics of these systems have closed-form expressions, but it is assumed that we do not have access to such expressions. We assume that we only have access to a set of system trajectories and learn a dynamics model from the state-action trajectories. \rev{The learned dynamics are represented as neural networks to the model-based control methods: deep contraction policy, MPC controller, and contraction-free learning. The RL implementation develops the policy directly from the same offline data set that is used to train the dynamics model in a model-free fashion.} This allows us to implement our algorithm on the learned systems while having an analytical baseline to compare against to quantify robustness. Additionally, we consider state and control sets $\X,\UU$ defined by box constraints in order to constrain the size of the training data. \rev{Clearly for such constraints, $\X$ is convex}. The dynamical systems we have chosen for our performance evaluation are as follows:
\begin{enumerate}[font=\bfseries]
  \item \textbf{2D Planar Car:} A planar vehicle that is capable of controlling its acceleration, $\alpha$, and angular velocity, $\omega$. Here $\x:=[p_x,p_y,\theta,v]$ and $\uu:=[\alpha,\omega]$ where $p_x,p_y$ are the planar positions, $v$ is the velocity, and $\theta$ is the heading angle. 
  The system dynamics are governed by: $\xdot = [v\cos(\theta), v\sin(\theta), \omega, \alpha]^\top$.
    
  \item \textbf{3D Drone:} An adaptation of a drone model that is given by~\cite{sun2020learning} and~\cite{7989693}. This model describes an aerial vehicle capable of directly controlling the rate of change of it's normalized thrust $\dot{F}$, and Euler Angles, $\dot{\phi}, \dot{\theta}, \dot{\psi}$. Here $\x:=[p_x,p_y,p_z,v_x,v_y,v_z,F,\phi,\theta,\psi]$ and $\uu:=[\dot{\phi},\dot{\theta},\dot{\psi}]$ where $p_i,v_i$ are the translational positions and velocities along the $i_{\text{th}}$ axis, respectively. Omitting the first order integrators in $p_i,F,\phi,\theta,\psi$ for brevity, the dynamics can then be expressed as $[\dot{v}_x , \dot{v}_y, \dot{v}_z] = [-F\sin(\theta), F\cos(\theta)\sin(\phi),g-\cos(\theta)\cos(\phi)]$,
 where $g$ is the acceleration due to gravity.
\end{enumerate}
For both systems we assume a timestep of $\Delta t = 0.1$s and a final time of $T = 10$s.

\subsection{Learning System Dynamics}
All of the continuous dynamical systems described above are represented to our controllers as fully connected neural networks which capture the discretization of the model integration: $\xtpl - \xt \approx \fest(\xt,\ut;\wf)$.
The training dataset $\D$ is generated by aggregating reference trajectories through the state space generated from an iLQR controller applied directly to the true dynamics $f(\xt, \ut)$. The reference trajectories $\Phi\left(\xt, \ut \right)$ were chosen such that $\xt \in \mathcal{X}$ and $\ut \in \mathcal{U}$ for all $t$. Trajectory data was used in order to implement a discounted multistep prediction error as in~\cite{Venkatraman2015ImprovingMP} until sufficient integration accuracy was achieved. 


\subsection{Controller Implementation}
The contraction metric and control policy neural networks, $\Test(\xt;\wT)$ and $\uest(\xt; \wU)$, are trained according to Algorithm~\ref{alg:deep_contraction}.

For our ablation study, we remove the contraction penalty term and simply find a policy for minimizing the tracking error norm. Without a contraction penalty, the impact of contraction conditions during the learning process vanishes. 
In order to create a controller for this case, each $\xt \in \mathcal{D}$ is forward evolved a number of time steps under the learned control policy and trained with a discounted cumulative loss of the tracking error norms over each timestep. 




For the MPC controller, the iLQR planner utilizes the learned dynamics model in order to calculate the linearization relative to the state and control inputs. This linearization is used along with weighting matrices $\vect{Q}=100\vect{I}, \vect{R}=1000\vect{I}$ in order to calculate an iLQR control law. 


In order to train an offline reinforcement learning algorithm like CQL, the algorithm needs access to state, action, and reward pairs. We reutilize the offline iLQR trajectories created for dynamics learning as training episodes for the offline CQL RL algorithm. The reward at each time step is taken to be the negative norm of the tracking error at the next time step given the currently taken action.

\begin{table*}[htbp]
  \setlength{\tabcolsep}{9pt}
  \caption{\rev{Tracking Error Norm RMSE, 3D Drone}}
  \begin{center}
  \begin{adjustbox}{width=\textwidth}
    \begin{tabular}{|c|c|c|c|c|c|c|c} \hline
      Dynamics Model & Test Loss & Contraction learning & No contraction term & MPC iLQR \\ \hline
      1 & 5.67e-05 & $\boldsymbol{1.905 \pm 0.651}$ & $2.391 \pm 0.968$ & $2.161 \pm 0.913$ \\
      2 & 8.19e-05 & $\boldsymbol{2.026 \pm 0.676}$ & $2.528 \pm 0.880$ & $2.966 \pm 1.593$ \\
      3 & 1.14e-04 & $\boldsymbol{2.214 \pm 0.889}$ & $3.315 \pm 0.917$ & $6.458 \pm 2.284$ \\
      4 & 1.58e-04 & $\boldsymbol{2.891 \pm 1.061}$ & $5.392 \pm 1.290$ & $\mathrm{N/A}^*$ \\
      5 & 2.64e-04 & $\boldsymbol{3.571 \pm 1.252}$ & $\mathrm{N/A}^*$  & $\mathrm{N/A}^*$ \\\hline
      \multicolumn{5}{l}{\rev{$^{\mathrm{*}}$Values of N/A represent cases where sufficiently stabilizing controllers were not generated.}}
    \end{tabular}
    \end{adjustbox}
    \label{tab::model_mismatch}
  \end{center}
\end{table*}

\begin{figure*}
    \centering
    \includegraphics[width=1\textwidth]{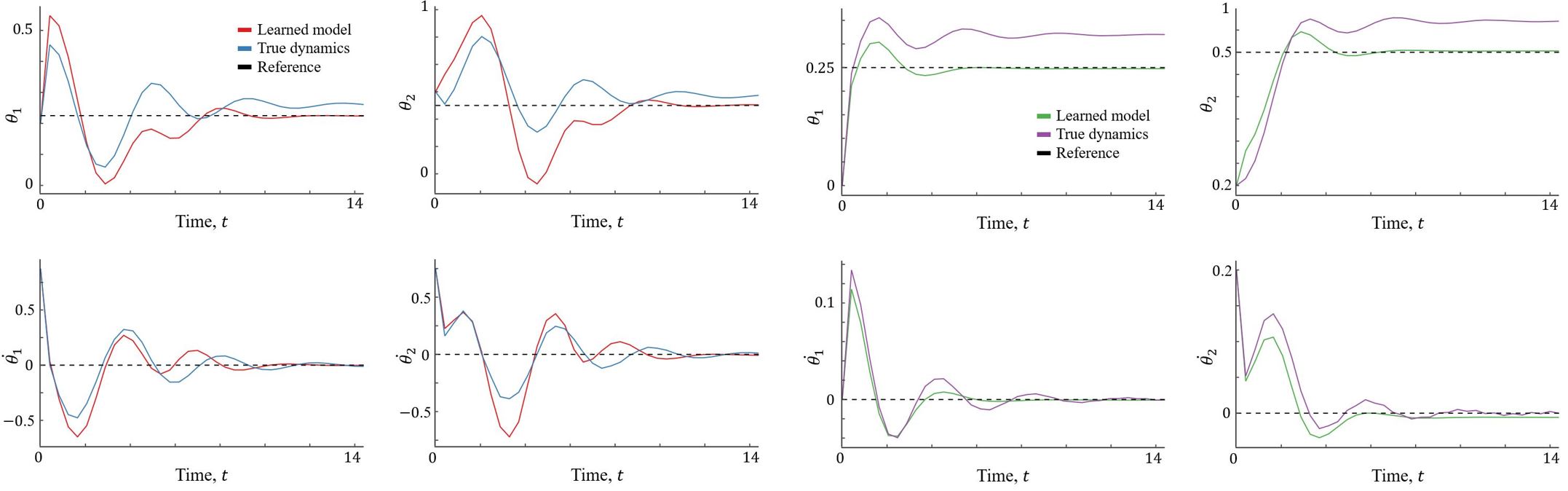}
    \caption{\small \rev{Angular position and angular velocity of the double pendulum system. The controlled learned model and controlled true dynamics are shown. Results are shown for two different sets of initial conditions}}
    \label{fig::pend}
\end{figure*}

\subsection{Performance Results}



In order to compare the performance of our method with the alternative implementations outlined above, we propose a number of metrics to compare the controllers:
\rev{
\begin{itemize}
    \item The time evolution of the tracking error, to quantify the controllers' ability to converge to the desired reference $\xr$.
    \item The converged tracking error versus the initial tracking error, to quantify the controllers' ability to operate over the working space $\XTU$.
    \item Root Mean Square Error (RMSE) of the tracking error versus learned model loss, to quantify the controllers' ability to deal with model mismatch.
\end{itemize}

For all analyses, the controllers were each presented with an identical set of 256 initial conditions within $\X$. The control methods were implemented as described above in an attempt to drive these initial states to the desired reference $\xr$. The results were aggregated over the 256 initial conditions
for the 2D car and 3D drone.
}

In the time evolution analysis, desirable controllers have trajectories that quickly converge, have minimal tracking error norm, and have high convergence precision. \rev{Results directly comparing all of the controllers relative to this performance measure for the two dynamical systems are given in Figure~\ref{fig::multiplot} (left and middle).} The results show that over the two different systems and a multitude of initial conditions, the contraction learning policy performs well relative to the proposed comparison controllers. For the simpler dynamic system of the two, the 2D planar car, the results are comparable among all controllers but favor the contraction controller, while the more complex drone environment shows the clear benefits of our approach. The enforcement of contraction conditions forces nearby trajectories to converge to one another, and when near the reference point, this has the effect of reducing the norm of the tracking error further than the systems designed without contraction in mind. The contraction controller consistently has the lowest mean norm of the tracking error over all the sampled initial states. 

\rev{
Comparing the converged tacking error, in this case, the average of the final 10 timestamps of each trajectory, versus the initial tracking error gives insight into the performance of the controllers' over the entire state and control space $\X$ and $\UU$. Cases with a higher initial tracking error represent trajectories that start closer towards the boundary of our working space $\XTU$. 
Favorable controllers are ones in which the converged tracking error remains constant or grows slowly as the initial tracking error increases. Figure~\ref{fig::multiplot} (right) directly shows this comparison. The results here clearly show that the MPC controller and the learned policy without the contraction terms have difficulty as the initial state norm gets further from the desired reference. For the MPC controller, the poor performance is likely caused by not having expressive enough dynamics due to the repeated linearization process. The contraction-free policy shows good performance for small initial tracking errors but quickly degrades as this value grows. Such a control method acts extremely locally. Training a collection of states to converge to the reference without the additional contraction structure does not yield favorable stability properties. The CQL policy and deep contraction learning generate trajectories with minimal degradation as the tracking error increases, with the contraction learning method consistently having the  highest degree of performance.
}

\rev{
Finally, for the 3D drone scenario, the impact of the learned dynamics model quality on the model-based controllers' performance is studied. To this end, multiple models of different quality were learned from the same offline data set. Since the CQL policy is directly learned from the offline data and does not utilize the learned dynamics model, this method is omitted from this analysis. In this case, favorable controllers are ones in which the error grows slowly with increasing model inaccuracy. Comparison of the RMSE values of the tracking error norm over the length of the trajectories for the varying quality dynamics models are shown in Table~\ref{tab::model_mismatch}. 
The contraction learning model shows favorable performance as dynamics model mismatch increases due to the robustness properties discussed in Section~\ref{sec:robustness}. For particularly low-quality learned dynamics models we even see that the deep contraction policy is able to generate stabilizing controllers where the contraction-free policy and MPC controller fail to do so.

\subsection{Non-control Affine Analysis}
In order to quantify the ability of our deep contraction policy learning to generalize to more complex systems, we perform an illustrative analysis of our controller on the double pendulum model given in~\cite{TS-TO:06}. Such a system is chaotic with a non-affine control input. 
Figure~\ref{fig::pend} shows the comparison of two scenarios where the designed controller was implemented on both the learned model and the true dynamics. The trajectories show the controller is able to accomplish the task when applied to the learned model. However, when applied to the true dynamics, the controller positions the arms with a slight positional error while keeping the angular velocity at zero.}

\section{Conclusion and Future Work}\label{sec:conclusion}
In this paper, we established a framework for learning a converging control policy for an unknown system from offline data. We leveraged Contraction theory and proposed a data augmentation method for encoding the contraction conditions directly into the loss function. We jointly learned the control policy and its corresponding contraction metric. We compared our method with several state-of-the-art control algorithms and showed that our method provides faster convergence, a smaller tracking error, lower variance of trajectories. \rev{For our future work, we would like to extend the current work to develop the stochastic confidence bound of our control design approach design.}


\section{Appendix}\label{sec:appendix}
\rev{
\begin{proof}[Lemma 1]
We ground our error analysis on the training error of the tuples $(\xt,\ut) \in \D$ and propagate the error to the general state and control tuples $(\x,\uu) \in \XTU$.
\begin{align*}
    &\| f(\x,\uu) - \fest(\x,\uu) \| \leq \| f(\xt,\ut) - \fest(\xt,\ut) \| + \\
    &\| (f(\x,\uu)-\fest(\x,\uu)) - (f(\xt,\ut)-\fest(\xt,\ut)) \|  \leq \\
    & \| f(\xt,\ut) - \fest(\xt,\ut) \| + \Lffest \left\| \xu - \xupr \right\|.
\end{align*}
The first and the second inequalities are obtained by adding and subtracting the terms $f(\xt,\ut)$ and $\fest(\xt,\ut)$, and also using the norm and Lipschitz constant  properties. If we define $E(\xt,\ut,\x,\uu)$ as the right hand side of the second inequality, then $\underset{(\x,\uu) \in \XTU}{\textit{max}}\,\,\underset{(\xt,\ut) \in \D}{\textit{min}}\,\,E(\xt,\ut,\x,\uu) \leq \varepsilon$
where  $$\varepsilon = \underset{(\xt,\xtpl,\ut) \in \D}{\textit{max}} \left\| \xtpl - \fest(\xt,\ut) \right\|  + \Lffest \mathsf{D},$$ which concludes the proof.
\end{proof}
\begin{proof}[Proposition 2]
We want to derive a sufficient condition which ensures that contraction condition~\eqref{eq::cont_notlearned} holds for the true dynamics model.
Using the learned dynamics model, the left-hand side of~\eqref{eq::cont_notlearned} for $\xt \in \X$ can be bounded for the true dynamics as 
\begin{align*}
   &\| \Test(\xtpl)(\xtplpr - \xtpl) \| -\| \Test(\xt) \Dxt \| \leq \\ 
   &\| \Test(\xtplest)(\xtplprest - \xtplest) \| -\| \Test(\xt) \Dxt \| + \\
    &\| (\Test(\xtpl)-\Test(\xtplest))(\xtplprest - \xtplest) \| + \\
    &\| \Test(\xtplest)((\xtplpr-\xtplprest)-(\xtpl-\xtplest)) \|+ \\
    &\| (\Test(\xtpl)-\Test(\xtplest))((\xtplpr-\xtplprest)-(\xtpl-\xtplest)) \|,
\end{align*}
where  $\xtpl = f(\xt,\uest(\xt))$, $\xtplpr = f(\xtpr,\uest(\xtpr))$, $\xtplest = \fest(\xt,\uest(\xt))$, and $\xtplprest = \fest(\xtpr,\uest(\xtpr))$. The inequality holds due to addition and subtraction of proper terms and norm properties. The inequality can be further simplified using the Frobenius norm of the contraction metric $\Test$. Since, by assumption, the entries of the contraction metric are bounded by $\gamma$, we have $\| \Test(\x)\|_{F} \leq n\gamma$. Having an upper bound estimate of the Lipschitz constant of entries of the contraction metric $\LTij$ and recalling that $\|(\xtplprest - \xtplest) \| \leq \varepsilon$ from Lemma~\ref{lem::estimation_error_bound}, leads to the result $\| (\Test(\xtpl)-\Test(\xtplest))\|_{F} \leq \varepsilon \sqrt{\underset{ij}{\sum}\LTij^2}$. In addition, using the estimated Lipschitz constant $\Lffest$, we have that $\|((\xtplpr-\xtplprest)-(\xtpl-\xtplest))\| \leq \Lffest \left\|  \xuxpr-\xux \right\|$. Now, using the Lipschitz constant of $\ux$ as $\Lux$, we have that $\|((\xtplpr-\xtplprest)-(\xtpl-\xtplest))\| \leq \Lffest \lambda( 1 + \Lux)$. Finally, we can write the following inequality: 
\begin{align}\label{eq::final_inequlity}
    &\| \Test(\xtpl)(\xtplpr - \xtpl) \| -\| \Test(\xt) \Dxt \| \leq \nonumber\\ 
    &\| \Test(\xtplest)(\xtplprest - \xtplest) \| - \| \Test(\xt) \Dxt \|+\nonumber\\
    & \lambda (\varepsilon \tau \Lfestu + (\varepsilon \tau + n\gamma) \Lffest ( 1 + \Lux)),
\end{align}
where $\tau = \sqrt{\underset{ij}{\sum}\LTij^2}$. With Lipschitz constant $\LC$, we can derive an upper bound for $C_{\fest(\xt,\uest(\xt))}(\x,\Dxt)$, $\xt \in \X$ and $\Dxt \in \Delta_t$, such that $C_{\fest(\x,\uest(\x))}(\xt,\Dxt) < \zeta$
where $$\zeta = \underset{\x \in \mathcal{X}}{\textit{max}}\,\,\underset{\xt \in \D}{\textit{min}}\,\,  C_{\fest(\xt,\uest(\xt))}(\xt,\Dxt) + \LC \| \xt - \x \|$$. Finally by taking the expectation on Equation~\eqref{eq::final_inequlity}, we get 
\begin{align*}
    &\EX_{\Dxt}\big( \| \Test(\xtpl)(\xtplpr - \xtpl) \| -\| \Test(\xt) \Dxt \| \big)\leq \\
    & \zeta + \lambda (\varepsilon \tau \Lfestu + (\varepsilon \tau + n\gamma) \Lffest ( 1 + \Lux))
\end{align*}
which concludes the proof. 
\end{proof}
}


\begin{thebibliography}{10}

\bibitem{lohmiller1998contraction}
W.~Lohmiller and J.-J.~E. Slotine, ``On contraction analysis for non-linear
  systems,'' {\em Automatica}, vol.~34, no.~6, pp.~683--696, 1998.

\bibitem{levine2020offline}
S.~Levine, A.~Kumar, G.~Tucker, and J.~Fu, ``Offline reinforcement learning:
  Tutorial, review, and perspectives on open problems,'' {\em arXiv preprint
  arXiv:2005.01643}, 2020.

\bibitem{NMB-ST-NM-JJES-VS:20}
N.~Boffi, S.~Tu, N.~Matni, J.~Slotine, and V.~Sindhwani, ``Learning stability
  certificates from data,'' {\em arXiv preprint arXiv:2008.05952}, 2020.

\bibitem{HT-SJC-JJES:21}
H.~Tsukamoto, S.~Chung, and J.~Slotine, ``Contraction theory for nonlinear
  stability analysis and learning-based control: A tutorial overview,'' {\em
  Annual Reviews in Control}, 2021.

\bibitem{TH-AZ-PA-SL:18}
T.~Haarnoja, A.~Zhou, P.~Abbeel, and S.~Levine, ``Soft actor-critic: Off-policy
  maximum entropy deep reinforcement learning with a stochastic actor,'' in
  {\em International conference on machine learning}, pp.~1861--1870, PMLR,
  2018.

\bibitem{AK-AZ-GT-SL:20}
A.~Kumar, A.~Zhou, G.~Tucker, and S.~Levine, ``Conservative q-learning for
  offline reinforcement learning,'' {\em arXiv preprint arXiv:2006.04779},
  2020.

\bibitem{TY-GT-LY-SE-JYZ-SL-CF-TM:20}
T.~Yu, G.~Thomas, L.~Yu, S.~Ermon, J.~Zou, S.~Levine, C.~Finn, and T.~Ma,
  ``Mopo: Model-based offline policy optimization,'' {\em Advances in Neural
  Information Processing Systems}, vol.~33, pp.~14129--14142, 2020.

\bibitem{SL-AK-GT-JF:20}
S.~Levine, A.~Kumar, G.~Tucker, and J.~Fu, ``Offline reinforcement learning:
  Tutorial, review, and perspectives on open problems,'' {\em arXiv preprint
  arXiv:2005.01643}, 2020.

\bibitem{LK-MB-SL:19}
L.~Kaiser, M.~Babaeizadeh, P.~Milos, B.~Osinski, R.~Campbell, K.~Czechowski,
  D.~Erhan, C.~Finn, P.~Kozakowski, S.~Levine, {\em et~al.}, ``Model-based
  reinforcement learning for atari,'' {\em arXiv preprint arXiv:1903.00374},
  2019.

\bibitem{TMM-JB-MCJ:20}
T.~Moerland, J.~Broekens, and M.~Jonker, ``Model-based reinforcement learning:
  A survey,'' {\em arXiv preprint arXiv:2006.16712}, 2020.

\bibitem{sun2020learning}
D.~Sun, S.~Jha, and C.~Fan, ``Learning certified control using contraction
  metric,'' {\em arXiv preprint arXiv:2011.12569}, 2020.

\bibitem{berkenkamp2017safe}
F.~Berkenkamp, M.~Turchetta, A.~P. Schoellig, and A.~Krause, ``Safe model-based
  reinforcement learning with stability guarantees,'' {\em arXiv preprint
  arXiv:1705.08551}, 2017.

\bibitem{SV-SK-HL-FB-JW-SW-MJ-RV:19}
S.~Vaskov, S.~Kousik, H.~Larson, F.~Bu, J.~Ward, S.~Worrall,
  M.~Johnson-Roberson, and R.~Vasudevan, ``Towards provably not-at-fault
  control of autonomous robots in arbitrary dynamic environments,'' {\em arXiv
  preprint arXiv:1902.02851}, 2019.

\bibitem{RT:09}
R.~Tedrake, ``Lqr-trees: Feedback motion planning on sparse randomized trees,''
  2009.

\bibitem{AM-RT:17}
A.~Majumdar and R.~Tedrake, ``Funnel libraries for real-time robust feedback
  motion planning,'' {\em The International Journal of Robotics Research},
  vol.~36, no.~8, pp.~947--982, 2017.

\bibitem{SLH-MC-SH-SB-JFF-CJT:17}
S.~Herbert, M.~Chen, S.~Han, S.~Bansal, J.~Fisac, and C.~Tomlin, ``Fastrack: A
  modular framework for fast and guaranteed safe motion planning,'' in {\em
  2017 IEEE 56th Annual Conference on Decision and Control (CDC)},
  pp.~1517--1522, IEEE, 2017.

\bibitem{SB-MC-JFF-CJT:17}
S.~Bansal, M.~Chen, F.~J.F, and C.~Tomlin, ``Safe sequential path planning of
  multi-vehicle systems under presence of disturbances and imperfect
  information,'' in {\em American Control Conference}, 2017.

\bibitem{HKK-JWG:02}
H.~K. Khalil and J.~W. Grizzle, {\em Nonlinear systems}, vol.~3.
\newblock Prentice hall Upper Saddle River, NJ, 2002.

\bibitem{JC-FC-CJT-KS:20}
J.~Choi, F.~Casta{\~n}eda, C.~Tomlin, and K.~Sreenath, ``Reinforcement learning
  for safety-critical control under model uncertainty, using control lyapunov
  functions and control barrier functions,'' in {\em Robotics: Science and
  Systems}, 2020.

\bibitem{AJT-AS-YY-ADA:20}
A.~Taylor, A.~Singletary, Y.~Yue, and A.~Ames, ``Learning for safety-critical
  control with control barrier functions,'' {\em Proceedings of Machine
  Learning Research}, vol.~1, p.~12, 2020.

\bibitem{AMZ-AME-MEL-FS:21}
A.~Zaki, A.~El-Nagar, M.~El-Bardini, and F.~Soliman, ``Deep learning controller
  for nonlinear system based on lyapunov stability criterion,'' {\em Neural
  Computing and Applications}, vol.~33, no.~5, pp.~1515--1531, 2021.

\bibitem{SMR-FB-AK:18}
S.~Richards, F.~Berkenkamp, and A.~Krause, ``The lyapunov neural network:
  Adaptive stability certification for safe learning of dynamical systems,'' in
  {\em Conference on Robot Learning}, pp.~466--476, PMLR, 2018.

\bibitem{Ar-HH-LL-HZ-DVD-ST-NM:20}
A.~Robey, H.~Hu, L.~Lindemann, H.~Zhang, D.~Dimarogonas, S.~Tu, and N.~Matni,
  ``Learning control barrier functions from expert demonstrations,'' in {\em
  IEEE Conference on Decision and Control}, pp.~3717--3724, 2020.

\bibitem{SC-MF-MM-GJP-VMP:21}
S.~Chen, M.~Fazlyab, M.~Morari, G.~Pappas, and V.~Preciado, ``Learning lyapunov
  functions for hybrid systems,'' in {\em Proceedings of the 24th International
  Conference on Hybrid Systems: Computation and Control}, pp.~1--11, 2021.

\bibitem{tsukamoto2021learning}
H.~Tsukamoto and S.-J. Chung, ``Learning-based robust motion planning with
  guaranteed stability: A contraction theory approach,'' {\em IEEE Robotics and
  Automation Letters}, 2021.

\bibitem{SMK-AB:11}
S.~Khansari-Zadeh and A.~Billard, ``Learning stable nonlinear dynamical systems
  with gaussian mixture models,'' {\em Transactions on Robotics}, vol.~27,
  no.~5, pp.~943--957, 2011.

\bibitem{JU-SH:17}
J.~Umlauft and S.~Hirche, ``Learning stable stochastic nonlinear dynamical
  systems,'' in {\em International Conference on Machine Learning},
  pp.~3502--3510, PMLR, 2017.

\bibitem{SSV-VS-JJES-MP:18}
S.~Singh, V.~Sindhwani, J.~Slotine, and M.~Pavone, ``Learning stabilizable
  dynamical systems via control contraction metrics,'' {\em arXiv preprint
  arXiv:1808.00113}, 2018.

\bibitem{AJT-VDV-HML-YY-ADA:19}
A.~Taylor, V.~Dorobantu, H.~Le, Y.~Yue, and A.~Ames, ``Episodic learning with
  control lyapunov functions for uncertain robotic systems,'' in {\em
  International Conference on Intelligent Robots and Systems}, pp.~6878--6884,
  IEEE, 2019.

\bibitem{JZK-GM:19}
J.~Kolter and G.~Manek, ``Learning stable deep dynamics models,'' {\em Advances
  in Neural Information Processing Systems}, vol.~32, pp.~11128--11136, 2019.

\bibitem{WL-JJES:98}
W.~Lohmiller and J.-J.~E. J.~J.~E.~Slotine, ``On contraction analysis for
  non-linear systems,'' {\em Automatica}, vol.~34, no.~6, pp.~683--696, 1998.

\bibitem{GRW-BPZ:96}
G.~Wood and B.~Zhang, ``Estimation of the lipschitz constant of a function,''
  {\em Journal of Global Optimization}, vol.~8, no.~1, pp.~91--103, 1996.

\bibitem{CK-GC-NO-DB:21}
C.~Knuth, G.~Chou, N.~Ozay, and D.~Berenson, ``Planning with learned dynamics:
  Probabilistic guarantees on safety and reachability via lipschitz
  constants,'' {\em IEEE Robotics and Automation Letters}, vol.~6, no.~3,
  pp.~5129--5136, 2021.

\bibitem{6386025}
Y.~Tassa, T.~Erez, and E.~Todorov, ``Synthesis and stabilization of complex
  behaviors through online trajectory optimization,'' in {\em 2012 IEEE/RSJ
  International Conference on Intelligent Robots and Systems}, pp.~4906--4913,
  2012.

\bibitem{DBLP:journals/corr/abs-2006-04779}
A.~Kumar, A.~Zhou, G.~Tucker, and S.~Levine, ``Conservative q-learning for
  offline reinforcement learning,'' {\em CoRR}, vol.~abs/2006.04779, 2020.

\bibitem{7989693}
S.~Singh, A.~Majumdar, J.-J. Slotine, and M.~Pavone, ``Robust online motion
  planning via contraction theory and convex optimization,'' in {\em 2017 IEEE
  International Conference on Robotics and Automation (ICRA)}, pp.~5883--5890,
  2017.

\bibitem{Venkatraman2015ImprovingMP}
A.~Venkatraman, M.~Hebert, and J.~Bagnell, ``Improving multi-step prediction of
  learned time series models,'' in {\em AAAI}, 2015.

\bibitem{TS-TO:06}
T.~Stachowiak and T.~Okada, ``A numerical analysis of chaos in the double
  pendulum,'' {\em Chaos, Solitons \& Fractals}, vol.~29, no.~2, pp.~417--422,
  2006.

\end{thebibliography}
\end{document}